\documentclass[runningheads]{llncs}
\usepackage{booktabs} 
\usepackage[T1]{fontenc}
\usepackage{graphicx}
\usepackage{hyperref}
\usepackage{color}
\usepackage{amsmath}
\usepackage{multirow}
\usepackage{tikz}
\usetikzlibrary{positioning,arrows.meta,fit,backgrounds}

\tikzstyle{startstop} = [rectangle, rounded corners, minimum width=1.5cm, minimum height=0.5cm,text centered, draw=black, fill=red!30]
\tikzstyle{io} = [trapezium, trapezium left angle=70, trapezium right angle=110, minimum width=1cm, minimum height=0.5cm, text centered, draw=black, fill=blue!30]

\tikzstyle{process} = [rectangle, minimum width=3cm, minimum height=0.5cm, text centered, draw=black, fill=orange!30]
\tikzstyle{decision} = [diamond, minimum width=0.5cm, minimum height=0.1cm, text centered, draw=black, fill=green!30]
\tikzstyle{process2} = [rectangle, minimum width=1cm, minimum height=0.5cm, text centered, draw=black, fill=orange!30]
\tikzstyle{arrow} = [thick,->,>=stealth]
\usetikzlibrary{shapes,shadows,arrows,positioning,angles,quotes}
\tikzset{My Arrow Style/.style={single arrow, fill=black!15, anchor=base, align=center,text width=2.3cm}}
\tikzstyle{arrow} = [thick,->,>=stealth]
\usetikzlibrary{calc,trees,positioning,arrows,fit,shapes,calc}

\begin{document}
\title{Automated Testing of LLM-Based Post Hoc Explainers Using Model Checking as an Oracle}
\titlerunning{Automated Testing of LLM-Based Post Hoc Explainers}
\author{Dennis Gross\inst{1,2}\orcidID{0009-0001-5734-0538} \and Helge Spieker\inst{2}\orcidID{0000-0003-2494-4279}}

\authorrunning{D. Gross et al.}

\institute{Institut für Kommunikations- und Prüfungsforschung gGmbH \and Simula Research Laboratory, Oslo, Norway\\
Corresponding author: \email{dennis@artigo.ai}}
\maketitle
\begin{abstract}
Large language models (LLMs) are used as post hoc explainers of sequential decision-making policies, producing natural-language explanations of why an action was chosen.
However, LLMs often generate plausible but incorrect statements, and no existing approach systematically tests whether such explanations are faithful to the underlying environment.
Two classic software testing challenges stand in the way: there is no oracle for the correctness of an explanation, and the test inputs, natural language queries about a policy's behavior, lack the structure needed for systematic test case generation.
We address both. Probabilistic model checking provides the test oracle, computing exact reference results against which LLM answers are graded automatically. 
A taxonomy of post hoc query categories structures the input space around the environment-level facts from which policy explanations are composed; test cases generated from it are prioritized by question-specific diagnostic difficulty scores.
Across seven MDP environments, the testing separates three open-weight LLMs: a reasoning model passes 85\% of test cases, a mid-size model 70\%, and a 1B model falls below the random baseline, while prioritization surfaces significantly harder cases than random selection.
Our results indicate how trustworthy LLM-generated explanations are in model-free settings, where the same LLMs are used but no oracle exists to verify them.
\end{abstract}
\section{Introduction}
\emph{Sequential decision-making tasks} are at the core of applications in domains such as gaming~\cite{DBLP:conf/icaart/Gross23}, manufacturing~\cite{gross2023model}, and healthcare~\cite{gross2026formally}: an agent repeatedly observes the current state of its environment, selects an action, and thereby stochastically influences the successor state to reach a fixed objective.
Such tasks are formally modeled as \emph{Markov decision processes (MDPs)}~\cite{DBLP:books/wi/Puterman94}.
We focus on MDPs with finitely many states and actions, and on \emph{memoryless} policies, which base each decision solely on the current state rather than on the history of past states and actions.
The consequences of a single decision unfold over subsequent steps.
Policies are typically obtained automatically, e.g., via \emph{deep reinforcement learning (RL)}~\cite{mnih2015human}, so the rationale behind an individual choice is often not accessible to the human user: the learned policy is a neural network whose internal computations are not human-interpretable.
To address this, \emph{large language models (LLMs)} have recently been employed as \emph{post hoc} explainers that analyze a decision after the fact, rather than making the policy interpretable by design~\cite{DBLP:journals/corr/abs-2504-00125}: after the policy has made a decision, the LLM is prompted to produce a natural language explanation of why that action was chosen in the given state.

However, LLMs are known to produce plausible-sounding but incorrect statements~\cite{DBLP:journals/corr/abs-2511-12869}, which raises a software testing problem: before such explanations can be trusted, the LLM-based explainer must itself be tested.
Testing it is challenging for two reasons.
First, there is no \emph{test oracle}~\cite{barr2014oracle}: judging whether an explanation is correct requires knowing the true properties of the environment and policy.
Second, the input space is unstructured: it is unclear which kinds of explanation queries exist and which concrete queries are worth posing.
Consequently, while prior work has evaluated LLM-generated explanations~\cite{DBLP:journals/corr/abs-2310-05797,suntharamoorthy2026llms} and even used them to improve policy performance~\cite{gross2024enhancing}, no existing approach systematically tests whether an LLM actually understands the underlying environment and produces explanations faithful to it.

In this paper, we present \emph{an automated approach for testing LLM-based post hoc explainers of sequential decision-making policies}. We address the oracle problem via \emph{probabilistic model checking}~\cite{DBLP:books/daglib/0020348}: given a formal model of the environment and a property specified in a temporal logic such as \emph{PCTL}~\cite{hansson1994logic}, a model checker exhaustively analyzes the state space and computes, for instance, the exact probability of reaching a goal or violating a safety condition.
These results are exact and carry formal guarantees, and thus serve as a reliable oracle against which the LLM's answers are graded automatically.
To structure the test input space, \emph{we introduce a taxonomy of post hoc query categories}, which guides generating targeted test queries, for example, which actions in a state are optimal or whether a state is safety-critical, and shows, per category, where different LLMs succeed or fail.
These queries target the environment-level facts, such as optimality and safety, from which any post hoc explanation of a policy's decision is composed.
Since not every state is equally informative for every query, \emph{we prioritize test inputs by a question-specific diagnostic score}, such as the fraction of non-optimal actions, and select the highest-ranked states for testing.
Our approach requires a formal model of the environment, but this restriction applies only to testing, not to its implications: the formal model provides a measurement instrument for the LLM's reasoning capabilities.
Test results obtained in this controlled setting thereby indicate how trustworthy LLM-generated explanations are in model-free settings, where the same LLMs are applied but no oracle exists to verify them.

\section{Related Work}
Our work touches four research areas: explainability for sequential decision-making, LLMs as post hoc explainers, testing and evaluating LLMs, and model checking for AI systems.

\paragraph{Explainability for sequential decision-making.}
A large body of work on \emph{explainable reinforcement learning
(XRL)}~\cite{DBLP:journals/corr/abs-2507-12599} generates \emph{post hoc} explanations of learned
policies, e.g., through policy summarization~\cite{DBLP:journals/corr/abs-2510-19244},
saliency maps~\cite{DBLP:journals/ral/SamadiKDD24}, and reward decomposition~\cite{DBLP:conf/nips/LinZYQLY19},
and, most recently, by employing LLMs as post hoc
explainers~\cite{DBLP:journals/corr/abs-2504-00125}.
All these approaches
focus on producing explanations; in contrast, we test whether such an
explainer produces explanations that are faithful to the underlying
environment.

\paragraph{LLMs as post hoc explainers.}
A growing line of work employs LLMs as post hoc explainers of learned
policies, prompting them to produce natural language accounts of why an agent
chose a given action~\cite{DBLP:journals/corr/abs-2504-05625,DBLP:conf/icdl/LuZMGLW23,DBLP:journals/tnn/CaoZCSCLLZYL25,DBLP:journals/corr/abs-2509-04809,DBLP:journals/corr/abs-2502-12530}.
Across this line of work, explanations are judged only against approximate
references, such as human studies~\cite{zhang2023explaining} and LLM-as-a-judge~\cite{belouadah2025evaluating}.
In contrast, we grade explanations against an exact ground truth from probabilistic model checking, organized along a taxonomy of post hoc query categories.

\paragraph{Testing and evaluating LLMs.}
Beyond explanations, a broad literature tests LLMs directly, ranging from behavioral test suites~\cite{ribeiro2020beyond} and hallucination benchmarks~\cite{lin2022truthfulqa,DBLP:journals/csur/JiLFYSXIBMF23} to benchmarks with verifiable ground truth, such as PlanBench~\cite{DBLP:conf/nips/ValmeekamMHSK23}, which grades LLM plans via automated validators; the lack of reliable test oracles for ML systems is a well-known challenge~\cite{zhang2020machine,barr2014oracle}.
However, where an exact ground truth exists, it tests the LLM's own task performance, not the faithfulness of its explanations of another system's decisions.
We use probabilistic model checking as a test oracle for exactly this purpose.

\paragraph{Model checking for AI systems.}
Probabilistic model checking has been used to verify learned policies against PCTL properties by model checking the policy-induced Markov chain~\cite{bacci2022verified,DBLP:conf/setta/GrossJJP22,gross2024probabilistic}, also in combination with explainability methods~\cite{DBLP:conf/sac/GrossS25,ashok2021dtcontrol} and with LLMs, either generating counterfactuals for policy repair~\cite{DBLP:conf/pts/GrossS24} or acting as the policy under verification~\cite{DBLP:journals/corr/abs-2510-06756}.
In all these works, the object under verification is the policy.
In contrast, we use model checking as a test oracle: its exact results serve as ground truth against which we grade the faithfulness of LLM-generated explanations.

\section{Background}
We introduce probabilistic systems, probabilistic model checking, LLMs, and the software testing concepts used throughout the paper, together with the notation used in later sections.

\subsection{Probabilistic Systems}
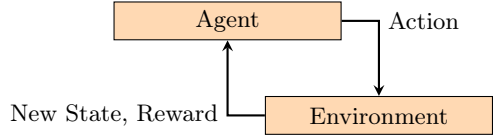
\begin{figure}[]
\centering
\scalebox{1}{
    \begin{tikzpicture}[]
     {};
    \node (agent1) [process] {Agent};
    \node (env) [process, below of=agent1,yshift=-0.25cm,xshift=2cm] {Environment};
    
    \draw [arrow] (agent1) -| node[anchor=west] {Action} (env);
    \draw [arrow] (env) -| node[anchor=east] {New State, Reward} (agent1);
    \end{tikzpicture}
}
\caption{A sequential decision-making system in which an agent interacts with an environment.
The agent receives a state and a reward from the environment based on its previous action. The agent then uses this information to select the next action via the policy, which it sends to the environment.}
\label{fig:rl}
\end{figure}

We model sequential decision-making (see Fig.~\ref{fig:rl}) as a \emph{Markov decision process (MDP)}, a tuple $\mathcal{M} = (S, s_0, \mathit{Act}, P, \mathit{AP}, L)$ where $S$ is a finite set of states, $s_0 \in S$ the initial state, $\mathit{Act}$ a finite set of actions, $P : S \times \mathit{Act} \times S \to [0,1]$ a transition function with $\sum_{s'} P(s, a, s') = 1$ for every enabled action $a$, $\mathit{AP}$ a finite set of atomic propositions, and $L : S \to 2^{\mathit{AP}}$ a labeling function mapping each state to the set of atomic propositions that hold in it.
We write $\mathit{Act}(s) \subseteq \mathit{Act}$ for the actions enabled in state $s$.
In each state, the agent picks an action $a$, and the environment draws the next state from the distribution $P(s, a, \cdot)$.
For example, in a slippery gridworld, the action \texttt{right} may reach the intended tile with probability $0.8$ and drift to a perpendicular neighbor otherwise.
Labels identify the states of interest: the \emph{goal} states are those labeled \textit{goal}, i.e.\ $\{s : \textit{goal} \in L(s)\}$, and the \emph{unsafe} set $U \subseteq S$ is defined analogously via an \textit{unsafe} label.
A \emph{(deterministic, memoryless) policy} $\pi : S \to \mathit{Act}$ chooses an action in each state; such policies suffice to attain the optimal reachability probabilities we consider.
Fixing $\pi$ resolves all choices and collapses the MDP into a \emph{induced discrete-time Markov chain (DTMC)}, whose behavior is fully determined by $P$ and $\pi$.

\subsection{Probabilistic Model Checking}
Given a specification written in \emph{Probabilistic Computation Tree Logic (PCTL)}~\cite{hansson1994logic}, a model checker computes the probability that the specification holds.
We use reachability properties of the form $\mathsf{P}_{\max}{=}?\ [\,\mathsf{F}\;\textit{goal}\,]$, read as ``what is the maximal probability of eventually ($\mathsf{F}$) reaching a \textit{goal} state?''; dually, $\mathsf{P}_{\min}{=}?\ [\,\mathsf{F}\;\textit{unsafe}\,]$ asks for the minimal such probability over all policies.
For every state $s$, the checker computes the \emph{property result} $V(s)$, the optimal probability of satisfying the property from $s$, and for every action $a \in \mathit{Act}(s)$ the \emph{action value} $Q(s,a)$, the property result obtained by taking $a$ in $s$ and acting optimally afterwards; we write $Q_{(1)} \ge Q_{(2)} \ge \dots$ for the decreasing sort of the action values in a state.
For the reachability properties $\mathsf{P}_{\max}[\mathsf{F}\,\textit{goal}]$ we consider, $V$ and $Q$ coincide exactly with the reinforcement learning value and action-value functions under the reward structure granting reward $1$ on entering a \textit{goal} state (made absorbing) and $0$ otherwise, undiscounted ($\gamma=1$), so the familiar $V$/$Q$ notation carries over.
We further use the \emph{danger} $D(s) = \mathsf{P}_{\min}[\,\mathsf{F}\,U\,]$, the minimal reachability probability of the unsafe set from $s$, obtained by a second model-checking run.
Tools such as Storm~\cite{DBLP:journals/sttt/HenselJKQV22} compute these quantities efficiently over the reachable~states.

\subsection{Large Language Models}
A \emph{large language model (LLM)} is a neural network based on the transformer architecture~\cite{DBLP:conf/nips/VaswaniSPUJGKP17}.
It operates on \emph{tokens}, i.e., elements of a finite vocabulary $\mathcal{V}$ into which a tokenizer segments natural-language text. Formally, an LLM defines a function $f_{\theta}\colon \mathcal{V}^{*} \to \Delta(\mathcal{V})$ that maps a token sequence to a probability distribution over the next token.
Text is generated \emph{autoregressively}: given an input sequence, the model samples a next token from $f_{\theta}$, appends it to the sequence, and repeats until a designated stop token is produced; the resulting token sequence is decoded back into text.

We use an LLM as a post hoc explainer as follows. A \emph{prompt} is a
natural-language input text that describes the environment, a state $s$, and
a query about the decision-making task in $s$. The prompt is tokenized and
passed to the model, which generates a natural-language response. A parser
then maps this response to a structured answer, such as a yes/no verdict or
an action ranking.

\subsection{Software Testing}
Testing executes a \emph{system under test (SUT)} on selected inputs, called \emph{test cases}, and compares the observed behavior against the expected behavior.
The mechanism that provides the expected behavior is the \emph{test oracle}~\cite{barr2014oracle}; its absence or unreliability, known as the \emph{oracle problem}~\cite{weyuker1982testing}, is a central challenge when testing systems whose correct output is not readily known, such as machine learning components~\cite{zhang2020machine}.
Since executing all possible test cases is generally infeasible, \emph{test case prioritization} orders them so that the most informative ones are executed first~\cite{rothermel2001prioritizing}.
In our setting, the SUT is the LLM-based explainer; a test case pairs the (extracted) MDP and its natural-language description with a state-level query; the model checker's exact results serve as the test oracle; and our diagnostic state ranking instantiates test-case prioritization.

\section{A Taxonomy of Post Hoc Test Queries}\label{sec:category}
The model checker provides a test oracle for each state: the property result $V(s)$ and the ranking of available actions by their values $Q(s, a)$.
Any post hoc explanation of a concrete policy decomposes into state-level claims about the environment, for instance, that an action is suboptimal, that a state is unsafe, or that a decision leads to a dead end.
Our queries test exactly these atomic claims against the \emph{optimal} behavior in the environment: a model that fails them cannot ground faithful explanations of any policy acting in it, so passing them is a necessary condition for explanation faithfulness.
Queries relative to a specific fixed policy $\pi$, whose induced DTMC the model checker analyzes in the same way, fit the same scheme.
A test case is a query posed to the LLM under test; we obtain verdicts by automatically comparing its answers against the oracle and prioritize test cases so that the most diagnostic ones are executed first (Section~\ref{sec:approach}).
Queries vary along three dimensions: \emph{object} (the property result or the action ranking), \emph{scope} (one state or a subset), and \emph{mode} (judging one object or comparing two).
The examples below illustrate each dimension but are not exhaustive; any query that can be answered by the oracle can be added in the same way.

\subsection{Derived Notions and Grading}\label{sec:metrics}
Beyond $V$, $Q$, and $D$ from the background, our queries use the following
derived notions. The \emph{optimal action set}
$A^\star(s) = \{a \in \mathit{Act}(s) : Q(s,a) = \max_{a'} Q(s,a')\}$
contains the maximizers of the action value in $s$, and the
\emph{worst action set} $A^\circ(s)$ is defined dually as the set of
minimizers. A state with $V(s) = 0$ is a \emph{dead end}: the property
can no longer be satisfied from it. A state $s$ is a \emph{bottleneck} iff
making $s$ absorbing reduces the maximal goal-reachability probability from
$s_0$ to $0$; this is decided exactly by re-checking the modified model.
Finally, $C_B(s)$ denotes the betweenness centrality~\cite{freeman1977} of
$s$ in the MDP's underlying directed graph, which has an edge $s \to s'$ iff
$P(s,a,s') > 0$ for some $a$.
$C_B$ serves only as a ranking heuristic, never as ground truth.
Grading is exact: the probability answered for the \emph{Satisfy} query is
correct iff it equals the model checker's result $V(s)$; a named best
(worst) action is correct iff it lies in $A^\star(s)$ ($A^\circ(s)$); and in
ranking queries, equal-valued actions may be ordered arbitrarily and are
graded as either-order-correct.

\subsection{Object}
The object is the quantity a query reads, giving two families.

\subsubsection{State queries}
Judgments about the state alone, independent of any action.
\begin{itemize}
    \item \emph{Satisfy property?} How likely the property is achieved from this state, answered as a probability estimate and graded against $V(s)$.
    \item \emph{Bottleneck?} Whether every successful run must pass through this state, i.e.\ the goal becomes unreachable without it.
\end{itemize}

\subsubsection{Preference queries}
Judgments about the actions at a given state.
\begin{itemize}
    \item \emph{Best action?} Which action is optimal, graded against $A^\star(s)$.
    \item \emph{Worst action?} Which action is most damaging, graded against $A^\circ(s)$.
    \item \emph{Full ranking?} How all actions order, graded against the sorted action values.
\end{itemize}
Note that a passing verdict on best-action test cases does not imply that the
LLM, deployed as a policy over full trajectories, reaches the target states.

\subsection{Scope}
A \emph{local} query reads the object at one state; a \emph{global} query reads it over a subset $S' \subseteq S$.
\begin{itemize}
    \item \emph{Bottleneck in the subset?} Whether any state in $S'$ is a bottleneck state.
    \item \emph{Dead ends in the subset?} Whether $S'$ contains a dead-end state, i.e.\ one with $V(s) = 0$.
\end{itemize}

\subsection{Mode}
An \emph{individual} query judges one object; a \emph{relational} query
compares two.
\begin{itemize}
    \item \emph{Which state is more promising?} Which of two states has the higher property result.
    \item \emph{Which state is safer?} Which of two states has the lower danger $D$.
    \item \emph{Which state is the bottleneck?} Given two states, which one every successful run must pass through.
\end{itemize}

\section{Testing Approach}\label{sec:approach}
Our approach takes as input an MDP, the PCTL properties of interest (the main property and, where danger queries are used, the safety property defining $D$), the LLMs under test, a query category from Section~\ref{sec:category} with a prompt template and a diagnostic scoring method, a \emph{test budget} (how many test cases to execute), and a \emph{sample size} (how many times each test case is repeated to account for LLM nondeterminism).
It outputs per-category verdicts for each LLM. 
The pipeline has four stages.

\paragraph{Oracle construction.} The MDP and the PCTL properties are
model-checked, yielding for every reachable state the property result $V(s)$,
the action values $Q(s,a)$, and, where required, the danger $D(s)$. From this we derive the expected answers the test cases need: the action ranking per
state, the optimal policy, the dead ends ($V(s)=0$), and the bottleneck
states, the latter by re-checking reachability with the candidate state made
absorbing (Section~\ref{sec:metrics}). For the best-action query, for
example, this yields each state's optimal action set $A^\star(s)$, the expected output against which the LLM's answer is later compared.

\paragraph{Test case generation and prioritization.} Each test case concerns a
single state (state and preference queries), a pair of states (mode queries), or a
subset of states (scope queries). The system generates candidate test cases of the
right kind, scores each by a diagnostic difficulty $\delta$, and keeps the
highest-$\delta$ (hardest) ones up to the test budget; the three binary categories
(bottleneck, subset-bottleneck, subset-dead-end) first balance positive and negative
cases and then rank within each class by $\delta$. Diagnostic difficulty targets the
cases most likely to be answered wrongly, via one of three notions: \emph{ambiguity}
(candidate values nearly tied, so the answer is barely determined), \emph{selectivity}
(a single correct choice among many distractors), and \emph{salience} (a decoy that
mimics the true structure). Using the notation of Sections~\ref{sec:category}
and~\ref{sec:metrics}, the difficulty is:
\begin{itemize}
  \item \emph{Satisfy} (ambiguity): $\delta=1-2\,|V(s)-\tfrac12|$, the property result closest to $\tfrac12$.
  \item \emph{Best action} (selectivity): $\delta=1-|A^\star(s)|/|\mathit{Act}(s)|$; few actions are optimal, so the best must be singled out from many distractors.
  \item \emph{Worst action} (selectivity): $\delta=1-|A^\circ(s)|/|\mathit{Act}(s)|$, symmetric to best action: few actions are worst.
  \item \emph{Full ranking} (ambiguity): $\delta=1-\min_i\,(Q_{(i)}-Q_{(i+1)})$, the smallest adjacent gap in the true ordering; equal-valued actions are graded as either-order-correct (Section~\ref{sec:metrics}).
  \item \emph{More promising} (ambiguity): $\delta=1-|V(s_1)-V(s_2)|$, the two states are close in property result.
  \item \emph{Safer state} (ambiguity): $\delta=1-|D(s_1)-D(s_2)|$, the two states are close in danger.
  \item \emph{Bottleneck} (salience): $\delta=C_B(s)$; the state lies on many paths, so it is the most bottleneck-like. The verdict itself is exact (Section~\ref{sec:metrics}), and $C_B$ only ranks candidates.
  \item \emph{Which-is-bottleneck} (salience): $\delta=C_B(s_{\mathit{decoy}})$, a central non-bottleneck decoy paired with the true bottleneck.
  \item \emph{Subset-bottleneck} (salience): $\delta=\max_{s\in S'}C_B(s)$, the subset's most bottleneck-like member.
  \item \emph{Subset-dead-end} (ambiguity, negatives): $\delta=1-\min_{s\in S'}V(s)$, a subset whose lowest value is near zero; positive subsets (containing a dead end, so $\min_{s\in S'}V(s)=0$) tie at $\delta=1$ and are sampled uniformly.
\end{itemize}
The $1-\text{gap}$ forms above assume values in $[0,1]$, as for the reachability
probabilities $V,D$ used throughout our experiments; more generally $\delta$ increases
as the relevant gap shrinks, and for expected-reward properties the gaps are
normalized by the value range so that $\delta\in[0,1]$.

\paragraph{Test execution.} Each selected test case is rendered into a prompt
by the template, with its state, pair, or subset filled in, and sent to every
LLM under test, repeated \emph{sample size} times. The template is the only
environment-specific part: it fixes how a state is described in words and what the LLM is asked. For the full-ranking query, it might read:
\begin{quote}\small
\ttfamily
You are an agent in a $5\times5$ gridworld. Your goal is to reach the
exit at $(4,4)$ while avoiding the trap at $(2,3)$.\\
Current position: \{state\}.\\
Available actions: \{actions\}.\\
Order all available actions from best to worst. Respond only with a
JSON object of the form \{"ranking": ["<best action>", ...,\\"<worst
action>"]\}.
\end{quote}
At execution time, the placeholders are filled from the selected state, e.g.\
\texttt{\{state\}} $\to$ \texttt{(1,3)} and \texttt{\{actions\}} $\to$
\texttt{up, down, left, right}, and the LLM returns a structured reply, for
instance: \texttt{\{"ranking": ["right", "down", "up", "left"]\}}.

\paragraph{Verdict.} Each answer is parsed and compared automatically against the oracle, and verdicts are aggregated over repetitions and per LLM.
The comparison depends on the category: an answered probability is compared for equality with $V(s)$, and a named best (worst) action is matched against $A^\star(s)$ ($A^\circ(s)$), a predicted ranking is compared to the true ordering up to ties, a more-promising or safer-state answer is checked against the two property results, a claimed bottleneck is verified against the exact bottleneck set, and a subset verdict is checked against whether $S'$ actually contains a bottleneck or dead end.

\begin{figure}[t]
\centering
\resizebox{\textwidth}{!}{%
\begin{tikzpicture}[
  font=\small,
  box/.style={draw, rounded corners=2pt, align=center,
              minimum height=12mm, inner sep=4pt},
  input/.style={box, fill=black!6,  text width=23mm},
  oracle/.style={box, fill=blue!8,  text width=26mm},
  llm/.style={box, fill=orange!12, text width=26mm},
  verdictbox/.style={box, fill=green!10, text width=21mm, minimum height=18mm},
  grp/.style={draw, black!50, dashed, rounded corners=3pt, inner sep=6pt},
  grplab/.style={font=\small\itshape, text=black!70, fill=white, inner sep=2pt},
  arr/.style={-{Stealth[length=2.4mm]}, semithick},
  lab/.style={font=\small, inner sep=1.5pt},
]
 
\node[input]  (in)   at (0, 0)      {MDP $\mathcal{M}$\\[1pt] + PCTL property};
 
\node[oracle] (mc)   at (4.0, 1.2)  {Model checker\\ (Storm)};
\node[oracle] (gt)   at (8.0, 1.2)  {\textbf{Test oracle}\\ $V(s)$, $Q(s,a)$, $D(s)$};
 
\node[llm]    (taxo) at (4.0,-1.2)  {Query taxonomy\\ \footnotesize object / scope / mode};
\node[llm]    (gen)  at (8.0,-1.2)  {Test cases\\ \footnotesize prioritized by difficulty $\delta$};
\node[llm]    (sut)  at (12.0,-1.2) {\textbf{LLM under test}\\ \footnotesize (post hoc explainer)};
 
\node[verdictbox] (v) at (15.2, 0)  {\textbf{Verdicts}\\ \footnotesize per query category};
 
\draw[arr] (in.east) -- ++(0.55,0) |- (mc.west);
\draw[arr] (in.east) -- ++(0.55,0) |- (taxo.west);
\draw[arr] (mc)   -- (gt);
\draw[arr] (taxo) -- (gen);
\draw[arr] (gen)  -- node[lab, above]{prompt} (sut);
 
\draw[arr, dashed] (gt) -- node[lab, right]{values for $\delta$} (gen);
 
\draw[arr] (gt.east)  -| node[lab, above, pos=0.22]{expected answer} (v.north);
\draw[arr] (sut.east) -| node[lab, below, pos=0.30]{LLM answer}      (v.south);
 
\begin{scope}[on background layer]
  \node[grp, fit=(mc)(gt)]         (oraclegrp) {};
  \node[grp, fit=(taxo)(gen)(sut)] (testgrp)   {};
\end{scope}
\node[grplab] at (oraclegrp.north) {1.\ oracle construction};
\node[grplab] at (testgrp.south)   {2.\ test generation \& execution};
 
\end{tikzpicture}%
}
\caption{Overview of our testing approach. Model checking the MDP against the
PCTL property yields an exact test oracle (top). The query taxonomy structures
the input space; test cases are prioritized by a diagnostic difficulty score
$\delta$ derived from the oracle's values and rendered into prompts for the
LLM under test (bottom). Verdicts are obtained by automatically comparing the
LLM's answers against the oracle.}
\label{fig:overview}
\end{figure}
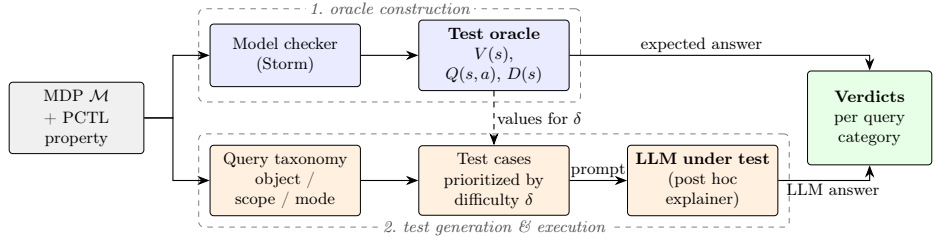

\section{Evaluation}\label{sec:eval}
We evaluate our approach to answer three research questions.
\begin{enumerate}
    \item[] \textbf{RQ1 (Prioritization):} Does the diagnostic ranking surface non-trivial cases?
    \item[] \textbf{RQ2 (Discrimination):} Do stronger LLMs pass more test cases than weaker ones?
    \item[] \textbf{RQ3 (Difficulty):} Do query categories differ in difficulty?
\end{enumerate}

\subsection{Experimental Setup}
\paragraph{Environments.} We use seven MDPs, each paired with a reachability
property whose model-checked result defines the oracle: \emph{Frozen Lake} (a
slippery $4\times4$ gridworld; reach the goal without falling into a hole;
one bottleneck), \emph{Wolf--Goat--Cabbage} (the river-crossing puzzle; move all
items across without an unsafe pair; four bottlenecks), \emph{Water Jug} (the
classic water-measuring puzzle; reach the target volume within a move budget;
no bottlenecks), \emph{Transporter} (a pickup-and-delivery task under stochastic
movement; collect and deliver the item to its destination; no bottlenecks),
\emph{Stock Market} (a trading task with stochastic price movements; reach a
target portfolio value without going bankrupt; no bottlenecks), \emph{Job Shop}
(a scheduling task with stochastic job durations; complete all jobs within a
deadline; no bottlenecks), and \emph{Dam} (a water-level control task with
stochastic inflow; keep the reservoir within safe bounds while meeting demand;
no bottlenecks). Environments without bottleneck states omit the three
bottleneck categories (reported as ``$-$'').

\paragraph{LLMs.} We test three open-weight models of increasing scale served
through Ollama: \emph{Gemma~3~1B}~\cite{gemmateam2025gemma3technicalreport} (small, run locally), \emph{Qwen3.5}~\cite{qwen35blog} (a larger
cloud model with a step-wise reasoning mode), and \emph{Gemma~4~31B}~\cite{gemmateam2026gemma4technicalreport}(a larger
cloud model).
A uniform-\emph{Random} answerer is the baseline; the optimal policy is the ceiling at $1$ by construction.

\paragraph{Protocol.}
All models are decoded at temperature~$0$ (greedy), at
which they are near-deterministic, so we use a sample size of $1$.
The test budget is $20$ states per category, selected by the diagnostic difficulty $\delta$ of Section~\ref{sec:approach}.
Answers are constrained to structured JSON, which is parsed automatically. 
The oracle is computed once per environment with the Storm model checker~\cite{DBLP:journals/sttt/HenselJKQV22} and reused across all models and categories.
Scores are the fraction of passing test cases in $[0,1]$ (higher is better).
We executed all experiments in a Docker container (16 GB RAM) on an AMD Ryzen 7 7735HS (16 threads) running Ubuntu 20.04.5 LTS.
For model checking, we use Storm 1.12.0.
The Ollama LLMs were hosted either on the same machine outside the Docker container or on the Ollama cloud and accessed via the REST API.

\subsection{Results}
\begin{figure}[t]
  \centering
  \resizebox{\columnwidth}{!}{\input{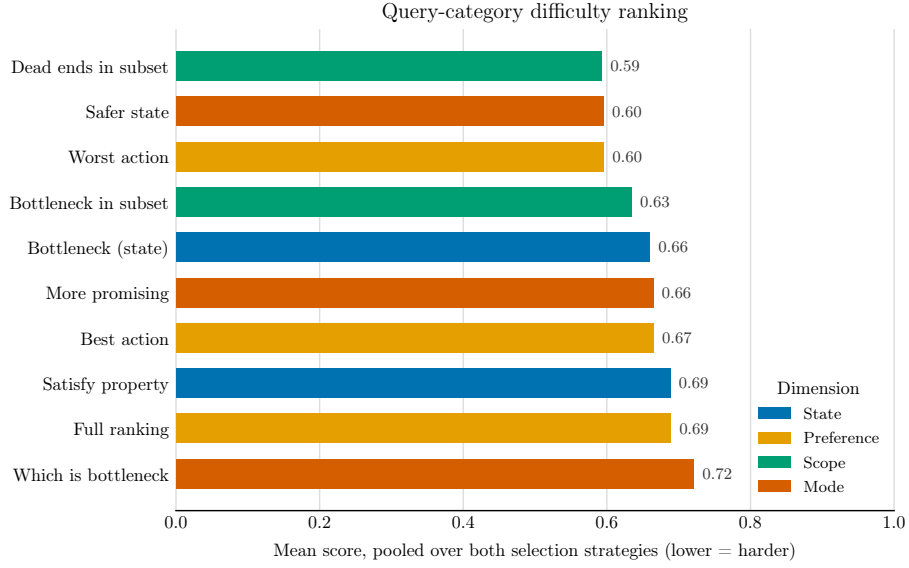}}
  \caption{Query-category difficulty, averaged over both selection
    strategies (diagnostic prioritization and random). Lower is harder;
    bars are colored by taxonomy dimension.}
  \label{fig:category-ranking}
\end{figure}

We first validate the selection method itself (RQ1), then read off LLM performance from the table (RQ2), and finally rank the categories by difficulty (RQ3).
Table~\ref{tab:llm-benchmark} reports per-category scores for all LLMs and environments, and Figure~\ref{fig:category-ranking} ranks the query categories by difficulty. 

\paragraph{Effect of diagnostic prioritization (RQ1).}
We compare, per category cell, the score under diagnostic prioritization against the score under random state selection; a \emph{lower} prioritized score means prioritization surfaced harder cases.
Over all $220$ comparable cells, prioritization yields the harder case in $75$ cells ($34.1\%$), the easier case in $61$ ($27.7\%$), and a tie in $84$ ($38.2\%$).
A one-sided Wilcoxon signed-rank test rejects the null in favor of prioritized $<$ random at $p = 0.035$.

Consistent with the mechanism, prioritization is \emph{not} expected to depress every model's score, and it does not.
It lowers a score only when a model is genuinely sensitive to oracle difficulty (small value gaps, action ambiguity, boundary states).
The \emph{Random} baseline is insensitive by construction; its expected score is fixed at the chance level of the answer space regardless of which states are chosen, so its prioritized-versus-random cells split roughly evenly, and its gaps are sampling noise.
The small \emph{Gemma 3 1B} is insensitive for a different reason: as the model comparison below shows, it is already at floor on most categories, so it has little room to fall further on cases the oracle labels difficult.
This answers RQ1: under diagnostic prioritization, we obtain more non-trivial test cases than with random test prioritization.

\paragraph{LLM performance (RQ2).}
Averaged over all environments, the reasoning model \emph{Qwen3.5} is
strongest at $0.85$, followed by \emph{Gemma 4 31B} at $0.70$, both
clearly above the \emph{Random} baseline at $0.51$.
Under diagnostic prioritization, the small \emph{Gemma 3 1B} sits \emph{below} random at $0.43$.
At $1$B parameters, it does not merely guess but commits to wrong answers in categories where random guessing would score at chance.
This deficit is specific to the prioritized test
cases, which surface exactly these hard states. Under random selection, the same model rises to $0.55$, essentially tied with random ($0.54$), so it is only exposed once the benchmark concentrates on diagnostic queries.
This answers RQ2: stronger models pass more test cases than both weaker models and the random baseline.

\paragraph{Category difficulty (RQ3).}
Figure~\ref{fig:category-ranking} ranks the query categories by the mean score pooled over all models and both selection strategies, so the ranking reflects intrinsic task difficulty rather than an artifact of which states were sampled.
The hardest categories are \emph{dead ends in subset} ($0.59$) and \emph{worst action} ($0.60$), followed by \emph{bottleneck in subset} ($0.63$); the easiest are the binary bottleneck queries, at $0.72$ for the relational \emph{which-is-bottleneck} variant.
This answers RQ3: the categories differ in difficulty.

\begin{table*}[t]
\centering
\caption{LLM performance across environments and query categories. State queries (satisfy property, bottleneck) and preference queries (best action, worst action, full ranking) vary along the object dimension; scope queries (bottleneck in subset, dead ends in subset) and mode queries (more promising, safer state, bottleneck) cover the scope and mode dimensions. Cells report the per-subcategory score in $[0,1]$; higher is better. Each cell reports the diagnostic-prioritization score followed by the random-selection score (prioritized/random); the prioritized score is shown in \textbf{bold} where prioritization yields a harder (lower) score than random selection.}
\label{tab:llm-benchmark}
\setlength{\tabcolsep}{4pt}
\renewcommand{\arraystretch}{1.2}
\resizebox{\textwidth}{!}{%
\begin{tabular}{ll *{2}{c} c *{3}{c} c *{2}{c} c *{3}{c} c c}
\toprule
& &
\multicolumn{2}{c}{\textbf{State}} & &
\multicolumn{3}{c}{\textbf{Preference}} & &
\multicolumn{2}{c}{\textbf{Scope}} & &
\multicolumn{3}{c}{\textbf{Mode}} & & \\
\cmidrule(lr){3-4} \cmidrule(lr){6-8} \cmidrule(lr){10-11} \cmidrule(lr){13-15}
\textbf{Environment} & \textbf{Model} &
Sat. & Bttl. & &
Best & Worst & Rank & &
Sub.B & Sub.D & &
Prom. & Safe & Bttl. & &
\textbf{Avg.} \\
\midrule
\multirow{4}{*}{Frozen Lake}
 & Gemma 3 1B & 0.65/0.65 & 0.13/0.13 & & 0.27/0.27 & 0.40/0.40 & 0.50/0.50 & & 0.45/0.45 & 0.50/0.50 & & \textbf{0.12}/0.38 & 0.50/0.50 & 0.67/0.20 & & 0.42/0.40 \\
 & Qwen3.5 & \textbf{0.74}/0.77 & 1.00/1.00 & & \textbf{0.55}/0.73 & \textbf{0.70}/0.90 & 0.93/0.85 & & \textbf{0.95}/1.00 & \textbf{0.95}/1.00 & & 0.88/0.88 & 0.50/0.50 & 1.00/1.00 & & \textbf{0.82}/0.86 \\
 & Gemma 4 31B & 0.83/0.83 & 0.93/0.93 & & 0.55/0.55 & 0.90/0.90 & 0.87/0.77 & & 0.55/0.50 & 1.00/0.90 & & \textbf{0.75}/0.88 & 0.33/0.33 & 0.93/0.93 & & 0.76/0.75 \\
 & Random & \textbf{0.48}/0.61 & \textbf{0.53}/0.67 & & \textbf{0.27}/0.36 & 0.20/0.20 & \textbf{0.42}/0.57 & & 0.40/0.40 & \textbf{0.40}/0.50 & & 0.62/0.50 & \textbf{0.33}/1.00 & 0.53/0.53 & & \textbf{0.42}/0.53 \\
\midrule
\multirow{4}{*}{Wolf--Goat--Cabbage}
 & Gemma 3 1B & 0.61/0.61 & 0.31/0.31 & & 0.78/0.78 & 0.20/0.20 & 0.57/0.57 & & 0.50/0.50 & 0.50/0.50 & & 1.00/0.00 & \textbf{0.00}/1.00 & \textbf{0.35}/0.55 & & \textbf{0.48}/0.50 \\
 & Qwen3.5 & 1.00/1.00 & 1.00/1.00 & & 1.00/1.00 & 1.00/1.00 & 1.00/1.00 & & \textbf{0.90}/1.00 & 1.00/1.00 & & 1.00/1.00 & 1.00/1.00 & 1.00/1.00 & & \textbf{0.99}/1.00 \\
 & Gemma 4 31B & 0.64/0.64 & 0.85/0.69 & & 1.00/1.00 & 0.40/0.40 & 0.87/0.87 & & \textbf{0.70}/0.80 & 0.50/0.50 & & 1.00/0.00 & \textbf{0.00}/1.00 & 0.95/0.95 & & 0.69/0.69 \\
 & Random & 0.45/0.45 & 0.54/0.54 & & \textbf{0.56}/0.67 & 0.60/0.40 & 0.77/0.67 & & \textbf{0.45}/0.60 & 0.60/0.60 & & 1.00/0.00 & 1.00/1.00 & \textbf{0.45}/0.50 & & 0.64/0.54 \\
\midrule
\multirow{4}{*}{Water Jug}
 & Gemma 3 1B & 0.57/0.41 & --/-- & & 0.35/0.35 & 0.71/0.71 & 0.27/0.27 & & --/-- & \textbf{0.50}/0.55 & & 0.00/0.00 & \textbf{0.00}/1.00 & --/-- & & \textbf{0.34}/0.47 \\
 & Qwen3.5 & 1.00/1.00 & --/-- & & 1.00/1.00 & 1.00/1.00 & \textbf{0.98}/1.00 & & --/-- & 0.90/0.80 & & 1.00/1.00 & 1.00/1.00 & --/-- & & 0.98/0.97 \\
 & Gemma 4 31B & \textbf{0.65}/0.75 & --/-- & & 0.70/0.70 & 0.82/0.82 & 0.85/0.83 & & --/-- & 0.70/0.55 & & 1.00/1.00 & 0.00/0.00 & --/-- & & 0.68/0.67 \\
 & Random & 0.48/0.41 & --/-- & & \textbf{0.40}/0.45 & \textbf{0.65}/0.71 & 0.73/0.64 & & --/-- & \textbf{0.40}/0.50 & & \textbf{0.00}/1.00 & 1.00/0.00 & --/-- & & \textbf{0.52}/0.53 \\
\midrule
\multirow{4}{*}{Transporter}
 & Gemma 3 1B & 0.99/0.69 & --/-- & & \textbf{0.80}/0.95 & \textbf{0.60}/0.70 & \textbf{0.50}/0.63 & & --/-- & 0.50/0.50 & & \textbf{0.00}/1.00 & \textbf{0.00}/1.00 & --/-- & & \textbf{0.48}/0.78 \\
 & Qwen3.5 & 1.00/0.85 & --/-- & & \textbf{0.75}/0.95 & 0.90/0.85 & 0.97/0.96 & & --/-- & 0.65/0.60 & & 1.00/1.00 & 1.00/1.00 & --/-- & & 0.90/0.89 \\
 & Gemma 4 31B & 1.00/0.80 & --/-- & & \textbf{0.90}/0.95 & \textbf{0.70}/0.75 & 0.91/0.86 & & --/-- & 0.75/0.50 & & 1.00/1.00 & \textbf{0.00}/1.00 & --/-- & & \textbf{0.75}/0.84 \\
 & Random & 0.46/0.40 & --/-- & & \textbf{0.20}/0.90 & \textbf{0.25}/0.50 & 0.73/0.72 & & --/-- & \textbf{0.35}/0.50 & & 1.00/1.00 & 1.00/0.00 & --/-- & & 0.57/0.57 \\
\midrule
\multirow{4}{*}{Stock Market}
 & Gemma 3 1B & 0.87/0.87 & --/-- & & 1.00/1.00 & 0.00/0.00 & 0.00/0.00 & & --/-- & 0.50/0.50 & & \textbf{0.00}/1.00 & \textbf{0.00}/1.00 & --/-- & & \textbf{0.34}/0.62 \\
 & Qwen3.5 & \textbf{0.53}/0.80 & --/-- & & 1.00/1.00 & 1.00/1.00 & 1.00/1.00 & & --/-- & 0.80/0.65 & & 1.00/1.00 & 1.00/1.00 & --/-- & & \textbf{0.90}/0.92 \\
 & Gemma 4 31B & \textbf{0.72}/0.87 & --/-- & & \textbf{0.80}/1.00 & 0.95/0.75 & \textbf{0.40}/0.50 & & --/-- & \textbf{0.60}/0.70 & & 1.00/1.00 & 1.00/1.00 & --/-- & & \textbf{0.78}/0.83 \\
 & Random & \textbf{0.46}/0.48 & --/-- & & \textbf{0.40}/1.00 & \textbf{0.30}/0.65 & 0.50/0.30 & & --/-- & 0.50/0.50 & & \textbf{0.00}/1.00 & 1.00/0.00 & --/-- & & \textbf{0.45}/0.56 \\
\midrule
\multirow{4}{*}{Job Shop}
 & Gemma 3 1B & 0.53/0.43 & --/-- & & 0.35/0.15 & \textbf{0.65}/0.80 & 0.47/0.25 & & --/-- & 0.50/0.50 & & 0.80/0.35 & 0.50/0.45 & --/-- & & 0.54/0.42 \\
 & Qwen3.5 & \textbf{0.69}/0.76 & --/-- & & \textbf{0.25}/0.60 & 0.70/0.40 & 0.77/0.62 & & --/-- & 0.50/0.50 & & 0.55/0.50 & 0.70/0.60 & --/-- & & 0.59/0.57 \\
 & Gemma 4 31B & \textbf{0.72}/0.79 & --/-- & & 0.50/0.30 & 0.80/0.00 & 0.58/0.57 & & --/-- & 0.50/0.50 & & 0.70/0.55 & 0.65/0.50 & --/-- & & 0.64/0.46 \\
 & Random & 0.75/0.54 & --/-- & & \textbf{0.30}/0.50 & \textbf{0.30}/0.60 & 0.60/0.55 & & --/-- & 0.65/0.50 & & 0.65/0.40 & 0.45/0.40 & --/-- & & 0.53/0.50 \\
\midrule
\multirow{4}{*}{Dam}
 & Gemma 3 1B & 0.51/0.33 & --/-- & & \textbf{0.00}/0.75 & \textbf{0.10}/0.65 & 0.92/0.75 & & --/-- & 0.50/0.50 & & \textbf{0.55}/0.75 & \textbf{0.45}/0.75 & --/-- & & \textbf{0.43}/0.64 \\
 & Qwen3.5 & \textbf{0.73}/0.87 & --/-- & & \textbf{0.95}/1.00 & 0.85/0.85 & 0.97/0.96 & & --/-- & 0.50/0.50 & & 0.70/0.60 & \textbf{0.45}/0.55 & --/-- & & \textbf{0.74}/0.76 \\
 & Gemma 4 31B & \textbf{0.63}/0.91 & --/-- & & \textbf{0.90}/1.00 & \textbf{0.25}/0.55 & 0.97/0.92 & & --/-- & \textbf{0.55}/0.65 & & \textbf{0.60}/0.70 & 0.50/0.45 & --/-- & & \textbf{0.63}/0.74 \\
 & Random & 0.73/0.61 & --/-- & & \textbf{0.20}/0.60 & \textbf{0.30}/0.45 & 0.72/0.68 & & --/-- & \textbf{0.35}/0.55 & & \textbf{0.35}/0.45 & \textbf{0.30}/0.65 & --/-- & & \textbf{0.42}/0.57 \\
\bottomrule
\end{tabular}%
}
\end{table*}

\section{Threats to Validity}
\paragraph{Internal.} Verdicts depend on
parsing model output into structured answers; we mitigate this with JSON-constrained decoding.
Greedy decoding with sample size $1$ captures the modal, not
average, behavior.

\paragraph{External.} Conclusions rest on seven (exactly model-checkable)
environments and three open-weight LLMs, and may not transfer to larger tasks or
proprietary models. Templates are handwritten and environment-specific; rewording the same environment can shift absolute scores.

\paragraph{Construct.} We test faithfulness to \emph{optimal} behavior; passingbest-action cases do not imply the LLM reaches the goal as a policy, nor that its wording is causally faithful to its own computation.
The difficulty $\delta$ (and betweenness centrality in particular) only ranks candidates; the verdicts remain exact.

\paragraph{Conclusion.} The prioritization effect is statistically significant over all $220$ cells ($p{=}0.035$) but modest in magnitude. 

\section{Conclusion}\label{sec:conclusion}
We presented an automated approach for testing LLM-based post hoc explainers: probabilistic model checking is an exact test oracle, a taxonomy of query categories structures the input space, and diagnostic scores prioritize test cases.
Across seven environments and three LLMs, the approach cleanly discriminates between models, localizes, per category, where each model can be trusted, and prioritizes significantly harder cases than random selection would.
Since the same LLMs are deployed where no oracle exists, these results quantify a trust gap that would otherwise go unmeasured.

Future work includes using failed test cases to fine-tune explainers~\cite{chen2025towards}.

\begin{credits}
\subsubsection{\ackname}
This work is funded by the European Union under grant agreement number 101091783 (MARS Project).

\subsubsection{\discintname}
The authors declare that they have no competing financial interests.
\end{credits}

%
%
\bibliographystyle{splncs04}
\bibliography{mybibliography}

@article{DBLP:journals/corr/abs-2310-05797,
  author       = {Nicholas Kroeger and
                  Dan Ley and
                  Satyapriya Krishna and
                  Chirag Agarwal and
                  Himabindu Lakkaraju},
  title        = {Are Large Language Models Post Hoc Explainers?},
  journal      = {CoRR},
  volume       = {abs/2310.05797},
  year         = {2023}
}

@article{weyuker1982testing,
  title={On testing non-testable programs},
  author={Weyuker, Elaine J},
  journal={The Computer Journal},
  volume={25},
  number={4},
  pages={465--470},
  year={1982},
  publisher={The British Computer Society}
}

@misc{gemmateam2025gemma3technicalreport,
      title={Gemma 3 Technical Report}, 
      author={Gemma Team},
      year={2025},
      eprint={2503.19786},
      archivePrefix={arXiv},
      primaryClass={cs.CL},
      url={https://arxiv.org/abs/2503.19786}, 
}

@misc{gemmateam2026gemma4technicalreport,
      title={Gemma 4 Technical Report}, 
      author={Gemma Team},
      year={2026},
      eprint={2607.02770},
      archivePrefix={arXiv},
      primaryClass={cs.CL},
      url={https://arxiv.org/abs/2607.02770}, 
}

@misc{qwen35blog,
    title = {Qwen3.5: Accelerating Productivity with Native Multimodal Agents},
    url = {https://qwen.ai/blog?id=qwen3.5},
    author = {Qwen Team},
    month = {February},
    year = {2026}
}

@inproceedings{chen2025towards,
  title={Towards consistent natural-language explanations via explanation-consistency finetuning},
  author={Chen, Yanda and Singh, Chandan and Liu, Xiaodong and Zuo, Simiao and Yu, Bin and He, He and Gao, Jianfeng},
  booktitle={Proceedings of the 31st International Conference on Computational Linguistics},
  pages={7558--7568},
  year={2025}
}

@inproceedings{ashok2021dtcontrol,
  title={dtControl 2.0: explainable strategy representation via decision tree learning steered by experts},
  author={Ashok, Pranav and Jackermeier, Mathias and K{\v{r}}et{\'\i}nsk{\`y}, Jan and Weinhuber, Christoph and Weininger, Maximilian and Yadav, Mayank},
  booktitle={International Conference on Tools and Algorithms for the Construction and Analysis of Systems},
  pages={326--345},
  year={2021},
  organization={Springer}
}

@inproceedings{bacci2022verified,
  title={Verified probabilistic policies for deep reinforcement learning},
  author={Bacci, Edoardo and Parker, David},
  booktitle={NASA Formal Methods Symposium},
  pages={193--212},
  year={2022},
  organization={Springer}
}

@inproceedings{DBLP:conf/nips/VaswaniSPUJGKP17,
  author       = {Ashish Vaswani and
                  Noam Shazeer and
                  Niki Parmar and
                  Jakob Uszkoreit and
                  Llion Jones and
                  Aidan N. Gomez and
                  Lukasz Kaiser and
                  Illia Polosukhin},
  title        = {Attention is All you Need},
  booktitle    = {{NIPS}},
  pages        = {5998--6008},
  year         = {2017}
}

@article{DBLP:journals/sttt/HenselJKQV22,
  author    = {Christian Hensel and
               Sebastian Junges and
               Joost{-}Pieter Katoen and
               Tim Quatmann and
               Matthias Volk},
  title     = {The probabilistic model checker {Storm}},
  journal   = {Int. J. Softw. Tools Technol. Transf.},
  volume    = {24},
  number    = {4},
  pages     = {589--610},
  year      = {2022}
}

@article{DBLP:journals/corr/abs-2510-06756,
  author       = {Dennis Gross and
                  Helge Spieker and
                  Arnaud Gotlieb},
  title        = {Verifying Memoryless Sequential Decision-making of Large Language
                  Models},
  journal      = {CoRR},
  volume       = {abs/2510.06756},
  year         = {2025}
}

@inproceedings{DBLP:conf/pts/GrossS24,
  author       = {Dennis Gross and
                  Helge Spieker},
  title        = {Enhancing {RL} Safety with Counterfactual {LLM} Reasoning},
  booktitle    = {{ICTSS}},
  series       = {Lecture Notes in Computer Science},
  volume       = {15383},
  pages        = {23--29},
  publisher    = {Springer},
  year         = {2024}
}

@inproceedings{DBLP:conf/sac/GrossS25,
  author       = {Dennis Gross and
                  Helge Spieker},
  title        = {{PCTL} Model Checking for Temporal {RL} Policy Safety Explanations},
  booktitle    = {{SAC}},
  pages        = {1514--1521},
  publisher    = {{ACM}},
  year         = {2025}
}

@article{gross2024probabilistic,
  title={Probabilistic model checking of stochastic reinforcement learning policies},
  author={Gross, Dennis and Spieker, Helge},
  journal={arXiv preprint arXiv:2403.18725},
  year={2024}
}

@inproceedings{DBLP:conf/setta/GrossJJP22,
  author       = {Dennis Gross and
                  Nils Jansen and
                  Sebastian Junges and
                  Guillermo A. P{\'{e}}rez},
  title        = {{COOL-MC:} {A} Comprehensive Tool for Reinforcement Learning and Model
                  Checking},
  booktitle    = {{SETTA}},
  series       = {Lecture Notes in Computer Science},
  volume       = {13649},
  pages        = {41--49},
  publisher    = {Springer},
  year         = {2022}
}

@article{barr2014oracle,
  title={The oracle problem in software testing: A survey},
  author={Barr, Earl T and Harman, Mark and McMinn, Phil and Shahbaz, Muzammil and Yoo, Shin},
  journal={IEEE transactions on software engineering},
  volume={41},
  number={5},
  year={2014}
}

@article{rothermel2001prioritizing,
  title={Prioritizing test cases for regression testing},
  author={Rothermel, Gregg and Untch, Roland H. and Chu, Chengyun and Harrold, Mary Jean},
  journal={IEEE Transactions on software engineering},
  volume={27},
  number={10},
  pages={929--948},
  year={2001},
  publisher={IEEE}
}

@article{zhang2020machine,
  title={Machine learning testing: Survey, landscapes and horizons},
  author={Zhang, Jie M and Harman, Mark and Ma, Lei and Liu, Yang},
  journal={IEEE Transactions on Software Engineering},
  volume={48},
  number={1},
  pages={1--36},
  year={2020},
  publisher={IEEE}
}

@article{DBLP:journals/csur/JiLFYSXIBMF23,
  author       = {Ziwei Ji and
                  Nayeon Lee and
                  Rita Frieske and
                  Tiezheng Yu and
                  Dan Su and
                  Yan Xu and
                  Etsuko Ishii and
                  Yejin Bang and
                  Andrea Madotto and
                  Pascale Fung},
  title        = {Survey of Hallucination in Natural Language Generation},
  journal      = {{ACM} Comput. Surv.},
  volume       = {55},
  number       = {12},
  pages        = {248:1--248:38},
  year         = {2023}
}

@inproceedings{lin2022truthfulqa,
  title={Truthfulqa: Measuring how models mimic human falsehoods},
  author={Lin, Stephanie and Hilton, Jacob and Evans, Owain},
  booktitle={Proceedings of the 60th annual meeting of the association for computational linguistics (volume 1: long papers)},
  pages={3214--3252},
  year={2022}
}

@inproceedings{DBLP:conf/nips/ValmeekamMHSK23,
  author       = {Karthik Valmeekam and
                  Matthew Marquez and
                  Alberto Olmo Hernandez and
                  Sarath Sreedharan and
                  Subbarao Kambhampati},
  title        = {PlanBench: An Extensible Benchmark for Evaluating Large Language Models
                  on Planning and Reasoning about Change},
  booktitle    = {NeurIPS},
  year         = {2023}
}

@inproceedings{ribeiro2020beyond,
  title={Beyond accuracy: Behavioral testing of NLP models with CheckList},
  author={Ribeiro, Marco Tulio and Wu, Tongshuang and Guestrin, Carlos and Singh, Sameer},
  booktitle={Proceedings of the 58th annual meeting of the association for computational linguistics},
  pages={4902--4912},
  year={2020}
}

@article{zhang2023explaining,
  title={Explaining agent behavior with large language models},
  author={Zhang, Xijia and Guo, Yue and Stepputtis, Simon and Sycara, Katia and Campbell, Joseph},
  journal={arXiv preprint arXiv:2309.10346},
  year={2023}
}

@article{belouadah2025evaluating,
  title={Evaluating the effectiveness of LLMs for explainable deep reinforcement learning},
  author={Belouadah, Ayoub and Ruiz-Rodr{\'\i}guez, Marcelo Luis and Kubler, Sylvain and Le Traon, Yves},
  journal={Machine Learning with Applications},
  pages={100795},
  year={2025},
  publisher={Elsevier}
}

@article{DBLP:journals/corr/abs-2502-12530,
  author       = {Xinyi Yang and
                  Liang Zeng and
                  Heng Dong and
                  Chao Yu and
                  Xiaoran Wu and
                  Huazhong Yang and
                  Yu Wang and
                  Milind Tambe and
                  Tonghan Wang},
  title        = {Policy-to-Language: Train LLMs to Explain Decisions with Flow-Matching
                  Generated Rewards},
  journal      = {CoRR},
  volume       = {abs/2502.12530},
  year         = {2025}
}

@article{DBLP:journals/corr/abs-2509-04809,
  author       = {Haechang Kim and
                  Hao Chen and
                  Can Li and
                  Jong Min Lee},
  title        = {TalkToAgent: {A} Human-centric Explanation of Reinforcement Learning
                  Agents with Large Language Models},
  journal      = {CoRR},
  volume       = {abs/2509.04809},
  year         = {2025}
}

@article{DBLP:journals/tnn/CaoZCSCLLZYL25,
  author       = {Yuji Cao and
                  Huan Zhao and
                  Yuheng Cheng and
                  Ting Shu and
                  Yue Chen and
                  Guolong Liu and
                  Gaoqi Liang and
                  Junhua Zhao and
                  Jinyue Yan and
                  Yun Li},
  title        = {Survey on Large Language Model-Enhanced Reinforcement Learning: Concept,
                  Taxonomy, and Methods},
  journal      = {{IEEE} Trans. Neural Networks Learn. Syst.},
  year         = {2025}
}

@inproceedings{DBLP:conf/icdl/LuZMGLW23,
  author       = {Wenhao Lu and
                  Xufeng Zhao and
                  Sven Magg and
                  Martin Gromniak and
                  Mengdi Li and
                  Stefan Wermter},
  title        = {A Closer Look at Reward Decomposition for High-Level Robotic Explanations},
  booktitle    = {{ICDL}},
  pages        = {429--436},
  publisher    = {{IEEE}},
  year         = {2023}
}

@article{DBLP:journals/corr/abs-2504-05625,
  author       = {Zhang Xi{-}Jia and
                  Yue Guo and
                  Shufei Chen and
                  Simon Stepputtis and
                  Matthew C. Gombolay and
                  Katia P. Sycara and
                  Joseph Campbell},
  title        = {Model-Agnostic Policy Explanations with Large Language Models},
  journal      = {CoRR},
  volume       = {abs/2504.05625},
  year         = {2025}
}

@book{DBLP:books/wi/Puterman94,
  author       = {Martin L. Puterman},
  title        = {Markov Decision Processes: Discrete Stochastic Dynamic Programming},
  series       = {Wiley Series in Probability and Statistics},
  publisher    = {Wiley},
  year         = {1994}
}

@inproceedings{DBLP:conf/nips/LinZYQLY19,
  author       = {Zichuan Lin and
                  Li Zhao and
                  Derek Yang and
                  Tao Qin and
                  Tie{-}Yan Liu and
                  Guangwen Yang},
  title        = {Distributional Reward Decomposition for Reinforcement Learning},
  booktitle    = {NeurIPS},
  pages        = {6212--6221},
  year         = {2019}
}

@article{DBLP:journals/ral/SamadiKDD24,
  author       = {Amir Samadi and
                  Konstantinos Koufos and
                  Kurt Debattista and
                  Mehrdad Dianati},
  title        = {{SAFE-RL:} Saliency-Aware Counterfactual Explainer for Deep Reinforcement
                  Learning Policies},
  journal      = {{IEEE} Robotics Autom. Lett.},
  volume       = {9},
  number       = {11},
  pages        = {9994--10001},
  year         = {2024}
}

@article{DBLP:journals/corr/abs-2510-19244,
  author       = {Yiyu Qian and
                  Su Nguyen and
                  Chao Chen and
                  Qinyue Zhou and
                  Liyuan Zhao},
  title        = {Interpret Policies in Deep Reinforcement Learning using {SILVER} with
                  RL-Guided Labeling: {A} Model-level Approach to High-dimensional and
                  Multi-action Environments},
  journal      = {CoRR},
  volume       = {abs/2510.19244},
  year         = {2025}
}

@article{DBLP:journals/corr/abs-2507-12599,
  author       = {L{\'{e}}o Sauli{\`{e}}res},
  title        = {A Survey of Explainable Reinforcement Learning: Targets, Methods and
                  Needs},
  journal      = {CoRR},
  volume       = {abs/2507.12599},
  year         = {2025}
}

@article{mnih2015human,
  author    = {Volodymyr Mnih and
               Koray Kavukcuoglu and
               David Silver and
               Andrei A. Rusu and
               Joel Veness and
               Marc G. Bellemare and
               Alex Graves and
               Martin A. Riedmiller and
               Andreas Fidjeland and
               Georg Ostrovski and
               Stig Petersen and
               Charles Beattie and
               Amir Sadik and
               Ioannis Antonoglou and
               Helen King and
               Dharshan Kumaran and
               Daan Wierstra and
               Shane Legg and
               Demis Hassabis},
  title     = {Human-level control through deep reinforcement learning},
  journal   = {Nat.},
  volume    = {518},
  number    = {7540},
  pages     = {529--533},
  year      = {2015}
}

@article{hansson1994logic,
  title={A logic for reasoning about time and reliability},
  author={Hansson, Hans and Jonsson, Bengt},
  journal={Formal aspects of computing},
  volume={6},
  number={5},
  pages={512--535},
  year={1994},
  publisher={Springer}
}

@book{DBLP:books/daglib/0020348,
  author    = {Christel Baier and
               Joost{-}Pieter Katoen},
  title     = {Principles of model checking},
  publisher = {{MIT} Press},
  year      = {2008}
}

@inproceedings{gross2023model,
  title={Model checking for adversarial multi-agent reinforcement learning with reactive defense methods},
  author={Gross, Dennis and Schmidl, Christoph and Jansen, Nils and P{\'e}rez, Guillermo A},
  booktitle={Proceedings of the International Conference on Automated Planning and Scheduling},
  volume={33},
  pages={162--170},
  year={2023}
}

@inproceedings{DBLP:conf/icaart/Gross23,
  author       = {Dennis Gross},
  title        = {Turn-Based Multi-Agent Reinforcement Learning Model Checking},
  booktitle    = {{ICAART} {(3)}},
  pages        = {980--987},
  publisher    = {{SCITEPRESS}},
  year         = {2023}
}

@article{gross2026formally,
  title={Formally Verifying and Explaining Sepsis Treatment Policies with COOL-MC},
  author={Gross, Dennis},
  journal={arXiv preprint arXiv:2602.14505},
  year={2026}
}

@article{DBLP:journals/corr/abs-2504-00125,
  author       = {Ahsan Bilal and
                  David Ebert and
                  Beiyu Lin},
  title        = {LLMs for Explainable {AI:} {A} Comprehensive Survey},
  journal      = {CoRR},
  volume       = {abs/2504.00125},
  year         = {2025}
}

@inproceedings{gross2024enhancing,
  title={Enhancing rl safety with counterfactual llm reasoning},
  author={Gross, Dennis and Spieker, Helge},
  booktitle={IFIP International Conference on Testing Software and Systems},
  pages={23--29},
  year={2024},
  organization={Springer}
}

@article{DBLP:journals/corr/abs-2511-12869,
  author       = {Muhammad Ahmed Mohsin and
                  Muhammad Umer and
                  Ahsan Bilal and
                  Zeeshan Memon and
                  Muhammad Ibtsaam Qadir and
                  Sagnik Bhattacharya and
                  Hassan Rizwan and
                  Abhiram Rao Gorle and
                  Maahe Zehra Kazmi and
                  Ayesha Mohsin and
                  Muhammad Usman Rafique and
                  Zihao He and
                  Pulkit Mehta and
                  Muhammad Ali Jamshed and
                  John M. Cioffi},
  title        = {On the Fundamental Limits of LLMs at Scale},
  journal      = {CoRR},
  volume       = {abs/2511.12869},
  year         = {2025}
}

@inproceedings{suntharamoorthy2026llms,
  author    = {Suntharamoorthy, Nivetha and Betten, J{\o}rn Eirik and Gross, Dennis and Valseth, Eirik and Spieker, Helge},
  title     = {Do LLMs Explain or Remember? Evaluating Language Model Explanations of Reinforcement Learning Failures},
  booktitle = {TRUST-AI: The Second European Workshop on Trustworthy AI},
  series = {IJCAI/ECAI 2026},
  year      = {2026},
}

@article{freeman1977,
  title={A set of measures of centrality based on betweenness},
  author={Freeman, Linton C},
  journal={Sociometry},
  pages={35--41},
  year={1977},
  publisher={JSTOR}
}

\end{document}